\pdfoutput=1

\documentclass[11pt]{article}
\usepackage[table]{xcolor}

\usepackage{acl}

\usepackage{times}
\usepackage{latexsym}
\usepackage{xspace}

\usepackage[T1]{fontenc}

\usepackage[utf8]{inputenc}

\usepackage{microtype}

\usepackage{inconsolata}

\usepackage[linesnumbered,ruled,vlined]{algorithm2e}
\usepackage{algpseudocode}
\usepackage{comment}

\usepackage{graphicx}
\usepackage{enumitem}
\usepackage[symbol]{footmisc}

\usepackage{amsmath}
\usepackage{amssymb}
\usepackage{url}
\usepackage{bbm}

\DeclareMathOperator*{\argmin}{arg\,min}

\usepackage{pifont}

\newcommand{\recoverr}{\texttt{ReCoVERR}\xspace}
\newcommand{\recoverrfull}{\textbf{Re}ason by \textbf{Co}llecting \textbf{V}isual \textbf{E}vidences that are \textbf{R}eliable and \textbf{R}elevant\xspace}
\newcommand{\recoverrshort}{\texttt{R}}
\newcommand{\dataset}{\mathcal{D}}
\newcommand{\model}[1]{\mathcal{M}_{#1}}
\newcommand{\modelfamily}[1]{\mathbb{M}_{#1}}
\newcommand{\system}[1]{\mathcal{S}_{#1}}

\newcommand{\prob}[1]{\pi_{\texttt{#1}}}
\newcommand{\answer}{a}
\newcommand{\question}{Q}
\newcommand{\evidences}[1]{\mathcal{E}_{#1}}
\newcommand{\evidence}[1]{e_{#1}}
\newcommand{\hypothesis}{\mathcal{H}}
\newcommand{\abstain}{\varnothing}
\newcommand{\selected}{S}
\newcommand{\vlm}{\texttt{VLM}}

\newcommand{\vision}{\texttt{Vis}}
\newcommand{\qas}{\texttt{QA}\rightarrow\texttt{S}}
\newcommand{\qgen}{\texttt{QGen}}
\newcommand{\nli}{\texttt{NLI}}
\newcommand{\rel}{\delta}

\newcommand{\coverage}{$\mathcal{C}$}
\newcommand{\risk}{$\mathcal{R}$}
\newcommand{\sprecall}{$\mathbb{R}_{\texttt{SP}}$}
\newcommand{\effrel}{$\Phi_1$}

\definecolor{darkgreen}{RGB}{0, 100, 0}
\definecolor{purple}{RGB}{128,0,128}
\definecolor{maroon}{RGB}{204,0,0}
\definecolor{navyblue}{RGB}{26,69,149}
\definecolor{Gray}{gray}{0.9}

\newcommand\result[2]{${\text{#1}}$} 
\newcommand\redresult[2]{{\color{maroon}#1}} 
\newcommand\blueresultwithstdev[2]{${\color{navyblue}{\textbf{#1}}_{\pm#2}}$} 
\newcommand\blueresult[2]{{$\color{navyblue}\textbf{#1}$}} 
\newcommand\qualexample[3]{#1 {\color{navyblue}\textbf{#2} [#3]}}

\SetKwComment{Comment}{// }{}
\SetKwInput{KwData}{Selective prediction hyperparameters}

\SetCommentSty{mycommfont}

\usepackage{multirow}

\usepackage{booktabs}
\usepackage{subfig}
\usepackage{todonotes}

\title{Selective “Selective Prediction”: \\Reducing Unnecessary Abstention in Vision-Language Reasoning}

\author{Tejas Srinivasan$^1$\thanks{\quad Work done by Tejas as part of an internship at AI2.}  \quad Jack Hessel$^2$ \quad Tanmay Gupta$^3$ \quad Bill Yuchen Lin$^3$ \\{\bf Yejin Choi$^{3,4}$ \quad Jesse Thomason$^1$ \quad Khyathi Raghavi Chandu$^3$} \\
  $^1$University of Southern California \quad
  $^2$Samaya AI \\
  $^3$Allen Institute for Artificial Intelligence \quad
  $^4$University of Washington \\
  \texttt{tejas.srinivasan@usc.edu}
  }
\begin{document}

\maketitle

\begin{abstract}
Selective prediction minimizes incorrect predictions from vision-language models (VLMs) by allowing them to abstain from answering when uncertain. 
However, when deploying a vision-language system with low tolerance for inaccurate predictions, selective prediction may be over-cautious and abstain too frequently, even on many correct predictions. 
We introduce \recoverr, an inference-time algorithm to reduce the over-abstention of a selective vision-language system without increasing the error rate of the system’s predictions. 
When the VLM makes a low-confidence prediction, instead of abstaining \recoverr tries to find relevant clues in the image that provide additional evidence for the prediction. 
\recoverr\ uses an LLM to pose related questions to the VLM, collects high-confidence evidences, and if enough evidence confirms the prediction the system makes a prediction instead of abstaining. 
\recoverr enables three VLMs (BLIP2, InstructBLIP and LLaVA-1.5) to answer up to 20\% more questions on the VQAv2 and A-OKVQA tasks without decreasing system accuracy, thus improving overall system reliability. 
Our code is available at \url{https://github.com/tejas1995/ReCoVERR}.
\end{abstract}

\section{Introduction}
\label{sec:intro}
\begin{figure}[t]
    \centering
    \includegraphics[width=\columnwidth]{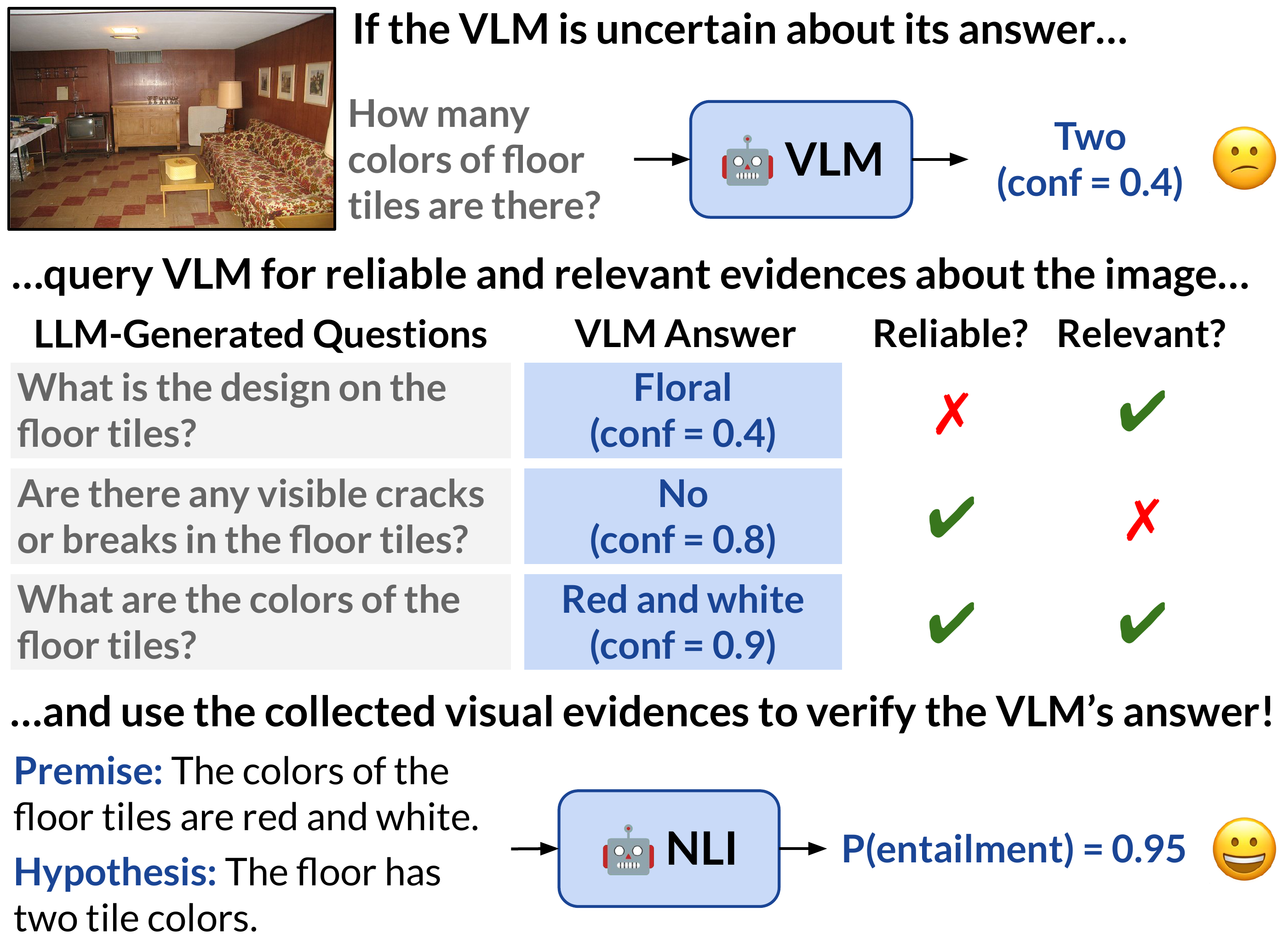}
    \caption{Illustration of \recoverr. 
    The VLM predicts that the floor has two tile colors with low confidence. 
    Instead of abstaining, \recoverr\ collects reliable and relevant visual evidences related to the question.
    \recoverr\ makes salient the evidence that the floor tiles are red and white, helping to verify the VLM's original answer.}
    \label{fig:teaser}
\end{figure}

Instruction-tuned vision-and-language models (VLMs)~\cite{dai2023instructblip, liu2023improved, laurencon2023obelics, bai2023qwen} have achieved strong accuracy on reasoning benchmarks, which typically require VLMs to produce an answer for each instance. 
For downstream use, however, these systems should abstain from answering when uncertain (e.g., by saying ``I don't know")~\cite{rajpurkar2018know}. 
Selective prediction systems~\cite{de2000reject, el2010foundations} aim to balance the number of predictions made (\textit{coverage}) and the error rate on predicted instances (\textit{risk}).

However, a vanilla selective prediction system with low tolerance for incorrect predictions will abstain too frequently to be practical, even when the model answer may be correct~\cite{whitehead2022reliable}.
For example, if a user specifies that the BLIP2~\cite{li2023blip} predictions should be right at least 90\% of the time, vanilla selective prediction will make a prediction for \textbf{just 4\%} of A-OKVQA~\cite{schwenk2022okvqa} questions, with 94\% of the correct predictions being abstained on.

We introduce \recoverr\ (\recoverrfull), an algorithm that increases the number of questions that a selective VLM system can answer confidently while adhering to a specified risk tolerance, \emph{without any additional training}. 
When the VLM is uncertain about its prediction for a given question, instead of abstaining outright, \recoverr tries to verify the prediction by recovering reliable supporting (or contradicting) evidence. 
This ability is predicated on two characteristics of VLMs. 
First, VLMs can produce well-calibrated confidence estimates (\S~\ref{subsubsec:vlm-calib}).
Second, VLMs can often correctly and confidently recover information in the image that entails a low-confidence initial prediction.
For instance, in Figure~\ref{fig:teaser}, when asked about the number of floor tile colors, a VLM (here, BLIP2) correctly answers ``two'' but with low confidence: simple threshold-based abstention would abstain on this instance.
But, when asked to identify the colors of the floor tiles, the model confidently answers ``red and white''. 
While it is obvious that ``red and white'' entails ``two colors'', the VLM was unable to confidently make that inference.
Based on these two insights, \recoverr searches for additional visual evidence by iteratively posing relevant questions to the VLM, and collecting answers as visual evidence if they are: a) \emph{reliable}, i.e, the VLM is highly confident in its answer, and b) \emph{relevant} to the question the VLM is trying to answer in the first place. 

We experiment on the VQAv2 and A-OKVQA visual reasoning tasks using three VLMs: BLIP2, InstructBLIP and LLaVA-1.5. 
For all three VLMs, \recoverr\ substantially increases the number of questions answered by the selective VLM system, while keeping system risk under the specified risk tolerance (Section~\ref{sec:results}). 
\recoverr\ is particularly helpful for BLIP2, which has not been trained on the target task, by improving coverage by 20\% and recall by 25-30\%. 
Our analysis reveals that the ability to give accurate confidence estimates, and high estimates for correct predictions, is crucial to \recoverr's performance. 
Further experiments demonstrate the importance of ensuring evidences are both reliable and relevant (Section~\ref{subsec:ablation}), and that \recoverr\ tuned for a single task can be directly applied to new tasks without further tuning (Section~\ref{subsec:task-generalization}). 
Our findings suggest that \recoverr\ is a promising solution towards building more reliable multimodal reasoning systems.

\section{Multimodal Selective Prediction}
\label{sec:background}
\begin{figure*}
    \centering
    \includegraphics[width=0.95\linewidth]{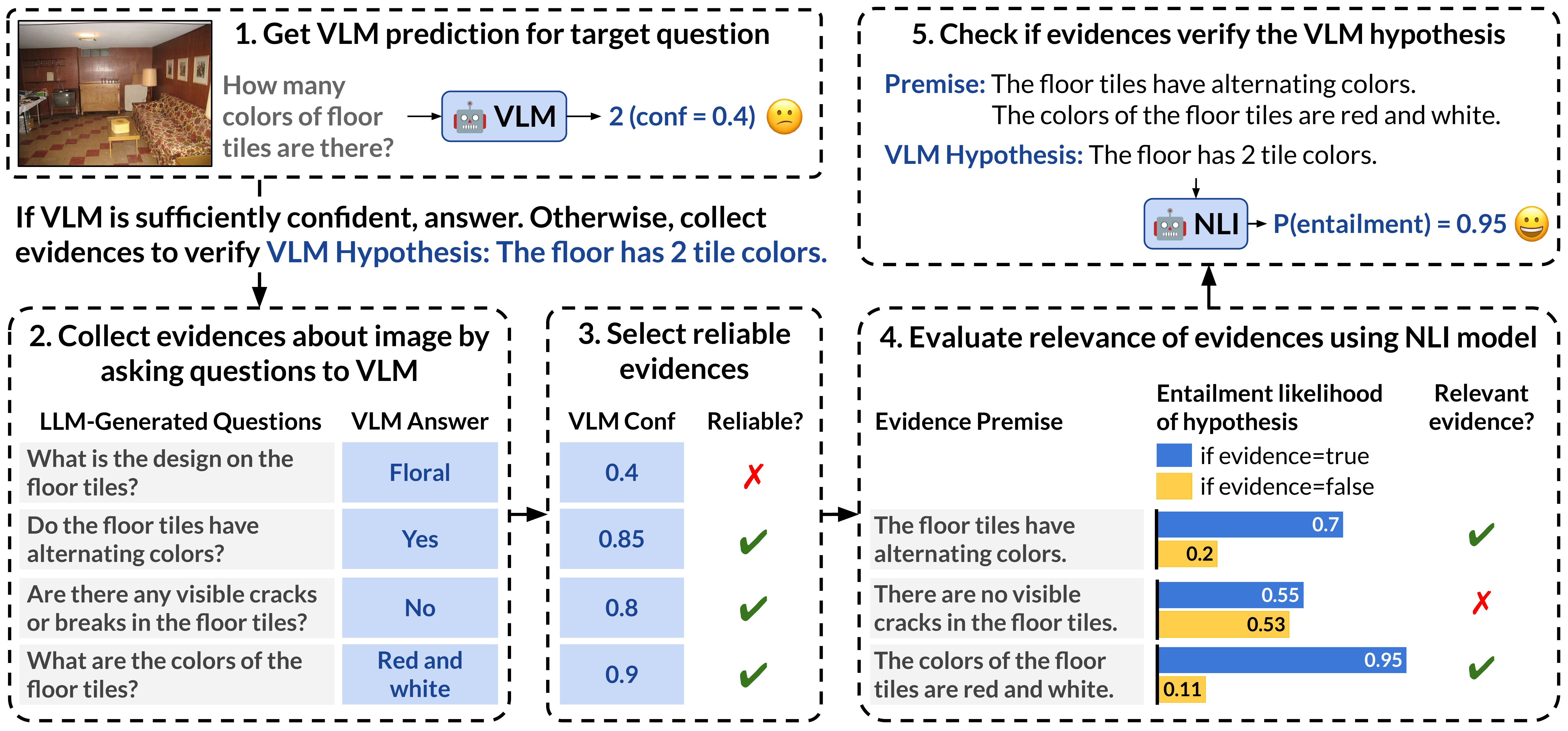}
    \caption{The \recoverr\ algorithm. If the VLM is uncertain in its prediction (1), \recoverr\ tries to verify the VLM hypothesis by collecting evidences. 
    \recoverr\ undertakes multiple turns of evidence collection, which involves generating visual evidences by using an LLM to ask questions to the VLM (2), retaining the reliable (3) and relevant (4) evidences, and checking whether the collected evidence entails the hypothesis (5).}
    \label{fig:recoverr}
\end{figure*}

We consider the task of answering a textual question about an image by drawing inferences from what is visually observed. 
A vision-language model (VLM) is given an input $x = (I, \question) \in \mathcal{X}$ consisting of an image $I$ and a question $\question$, and aims to produce an answer $\answer \in A$ from a closed or open set of possible answers. 

Different from popular benchmarks of this form which require models to make a prediction for each instance, we evaluate models on the selective prediction setting, where abstention on individual instances is allowed~\cite{de2000reject}.  
A decision function $g$, which has access to the image, question, VLM prediction, and associated confidence\footnote{For brevity, we note arguments for $g$ when relevant.} determines whether the system produces an answer or abstains (denoted by $\varnothing$). 
The selective VLM system $\system{\vlm}: \mathcal{X} \rightarrow \mathcal{A} \cup \{ \varnothing \}$ is defined as:
\begin{align*}
    \answer &= \model{\vlm}(x) \\
    \system{\vlm}(x) &= \begin{cases} 
        \answer, & \text{if } g(\answer) = 1 \\
        \varnothing, & \text{if not } g(\answer) = 0
    \end{cases}
\end{align*}

Prior work~\citep{whitehead2022reliable} employs confidence-based selection, where instead of assuming access to a VLM which has been trained to abstain with explicit supervision, we assume a VLM that can produce both an answer, as well as a confidence score for that answer
$\prob{\vlm}: \mathcal{A} \rightarrow [0,1]$
that estimates $\prob{\vlm}(\answer) = P(\answer | \question, I; \model{\vlm})$. 
Confidence-based selection relies on the confidence
$\prob{\vlm}(\answer)$ being higher for correct answers than incorrect ones, on average. 
The decision function $g$ is a simple thresholding function, with a threshold $\gamma$:  $g(\answer; \gamma) = \mathbbm{1}\{ \prob{\vlm}(\answer) \geq \gamma \}$.
We refer to this method as \textit{vanilla selective prediction}.

\subsection{Evaluating Selective Predictors}
We evaluate selective prediction systems via coverage and risk~\cite{el2010foundations}. 
Given a labeled evaluation dataset $\dataset = \{ (I_i, \question_i, \answer_i) \}_{i=1}^N$.
\emph{Coverage} (\coverage) is the percentage of questions in $\dataset$ where the system chooses to make a prediction.  
\emph{Risk} (\risk) is the error rate on the questions where a prediction is made. 
For a $\gamma$-threshold selective prediction system, these metrics are computed as:
\begin{gather}
    \mathcal{R}(\gamma) = \frac{\sum_{x_i \in \dataset} (1-\text{Acc}(a_i)) \cdot g(a_i; \gamma)}{\sum_{x_i \in \mathcal{D}} g(a_i; \gamma)} \label{eq:risk} \\
    \mathcal{C}(\gamma) = \frac{\sum_{x_i \in \dataset}g(\answer_i; \gamma) }{\mid \dataset \mid} \label{eq:coverage}
\end{gather}
where a lower $\gamma$ trades off increased coverage for increased risk. 
In practice, when deploying a selective VLM system with an application-specific risk tolerance $r \in [0, 1]$, 
the system designer would determine an appropriate setting for the hyperparameters of $g$ using a calibration set. For example, if $g$ is the simple $\gamma$-thresholding function, one would choose $\gamma_{@r}$ to maximize coverage while being below the risk threshold on the calibration set:
\begin{align*}
    \gamma_{@r} = \argmin_{\gamma \in [0, 1]} R(\gamma) \leq r
\end{align*}

\section{\recoverr}
\label{sec:recoverr}
In practice, reasonable risk tolerance thresholds often lead to untenable coverage, with many accurate predictions being discarded. A threshold-selective BLIP2~\cite{li2023blip} system, for example, answers fewer than 4\% of questions in A-OKVQA at 10\% risk, but these represent only 6\% of questions for which the model prediction was correct.

We introduce \recoverr\ (\recoverrfull), an algorithm that increases a selective VLM system's coverage while remaining under the given risk tolerance (Figure~\ref{fig:recoverr}). 
For each question $\question$ for which the VLM predicts an answer $\answer$ with $\prob{\vlm}(\answer) < \gamma_{@r}$, instead of abstaining, \recoverr\ uses large language models to generate follow-up questions about the image.
These questions are answered by the VLM, and the resulting QA-pair is added as \textit{evidence} if it is both \textit{reliable}---the VLM confidence in the answer is high---and \textit{relevant}---the introduction of the new evidence affects the downstream confidence in the hypothesized answer $\answer$ to $\question$. 
If sufficient evidences are collected to confidently entail the hypothesis, \recoverr\ elects to make a prediction instead of abstaining.

Let $\dataset_\abstain$ be the subset of test set $\dataset$ where the selective prediction system would have abstained, based on the confidence threshold $\gamma_{@r}$, and $\dataset_\selected$ be the set of questions that were answered. 
\begin{gather*}
    \dataset_\abstain = \{ x_i \in \dataset; \prob{\vlm}(\answer_i) < \gamma_{@r}\} ;
    \dataset_\selected = \dataset \backslash \dataset_\abstain
\end{gather*}
\recoverr\ aims to answer additional instances $\dataset_{\recoverrshort} \subseteq \dataset_{\abstain}$, while keeping the combined risk of $\dataset_\selected \cup \dataset_{\recoverrshort}$ under $r$. 
Since risk of $\dataset_\selected$ is already estimated to be $r$, the risk of $\dataset_{\recoverrshort}$ must also be $\leq r$. 
For each instance in $\dataset_\abstain$, \recoverr\ makes an online decision of whether to answer the question, independent of other instances in $\dataset_\abstain$. 
Therefore, the expected risk of any instance $x_i \in \dataset_{\recoverrshort}$ should be, at most, $r$; in other words, the likelihood of each instance in $\dataset_{\recoverrshort}$ being correct should be $\geq 1-r$. 
For example, if our system has a specified risk tolerance of 20\%, \recoverr\ aims to answer additional questions of which 80\% are correct.

\begin{algorithm}[t]
    \small
    \caption{\recoverr\ Pseudocode}
    \label{alg:recoverr-pseudocode}
    \KwData{
    \\$r \in [0, 1]$ : Risk tolerance of system \\
    $\gamma_{@r}$ : VLM conf threshold corresponding to risk $r$ \\
    \textbf{\recoverr\ Model-based Tools:} \\
    $\modelfamily{\vision}$: Set of additional vision models that output information about the image in text form\\
    $\model{\qgen}$: Question generation model \\
    $\model{\qas}$: LM to paraphrase QA pair into sentence \\
    $\model{\nli}$: NLI model \\
    \textbf{\recoverr\ hyperparameters:} \\
    $\rel_{\texttt{min}}.$ : minimum relevance of ``relevant'' evidences\\
    $\prob{NLImin}$: Minimum entailment confidence to answer \\
    $N$: Maximum number of turns of evidence collection\\
    $K$: Questions generated at each turn \\ 
    \textbf{Inputs:} Question $\question$, image $I$, VLM $\model{\vlm}$ \\ \quad }
    $\answer, \prob{\vlm}(\answer) \gets \model{\vlm}(I, \question$) \\
    \If{$\prob{\vlm}(\answer) \geq \gamma_{@r}$}{
        \Return $\answer$
    }
    $\hypothesis \gets \model{\qas}(\question, \answer)$ \\
    $\evidences{R}, \evidences{RR} \gets $\textsc{InitImageEvidences}($I; \modelfamily{\vision}$) \\
    \For{$i = 1$ \textbf{ to } $N$}{
        $e_{1...K} \gets $\textsc{CollectKEvidences}($I, \question, \evidences{R};$ \\
            \hspace{121pt} $\model{\qgen}, \model{\vlm}$) \\
        \For{$j = 1$ \textbf{ to } $K$}{
            \If{\textsc{VLMConfidence}($e_j ; \model{\vlm}$) $\geq 1-r$}{
                $\evidences{R} \gets \evidences{R} + [e_j]$ \\
                \If{\textsc{Relevance}($e_j ; \model{\nli}$) $\geq \rel_{\texttt{min}}.$}{
                    $\evidences{RR} \gets \evidences{RR} + [e_j]$
                }
            }
        }
        \If{$P(\evidences{RR} \text{ entails } \hypothesis; \model{\nli}) \geq \prob{NLIMin}$}{
            \Return $\answer$
        }
    }
    \Return $\varnothing$
\end{algorithm}
Algorithm~\ref{alg:recoverr-pseudocode} describes \recoverr in detail, including all hyperparameters. 
Given an image $I$ and a question $\question$, \recoverr\ begins by generating an answer $\answer$ using the VLM $\model{\vlm}$. The VLM also returns a confidence score $\prob{VLM}(\answer) \in [0, 1]$. 
If the confidence is higher than our confidence threshold for selective prediction $\gamma_{@r}$, we can simply return the answer. 
If not, we use a model $\model{\qas}$ that converts the question-answer pair $(\question, \answer)$ into a hypothesis statement $\hypothesis$. 
\recoverr\ will now try to verify this hypothesis by collecting reliable and relevant visual evidences. 

\paragraph{Initialize Evidences from Vision Tools: }
We first gather some general information about the image by invoking a set of vision models, $\modelfamily{\vision}$, that capture visual information in language form (the specific tools we use in our instantiation of \recoverr\ are highlighted in Section~\ref{subsec:implementation}). 
Information from these tools instantiate two evidence sets: $\evidences{R}$, a set of reliable evidences about the image, and $\evidences{RR}$, a set of reliable \emph{and} relevant evidences. 

\recoverr\ performs up to $N$ turns of evidence collection. 
In each turn, $K$ new visual evidences are generated from the VLM, the reliable and relevant ones are retained, and we check whether the collected evidences sufficiently entail the hypothesis. 
If we are unable to verify the hypothesis at the end of $N$ turns, we abstain from answering.

\paragraph{Generating Visual Evidences:} 
We prompt a question generation model $\model{\qgen}$ to generate a set of $K$ sub-questions $q_{1...K}$, conditioned on the target question $\question$ and the already-collected reliable evidences $\evidences{R}$. 
For each sub-question $q_j, j \in \{1...K\}$, the VLM produces an answer $a_j$ with confidence $\prob{\vlm}(a_j)$. 
The paraphraser model $\model{\qas}$ paraphrases the pair $(q_j, a_j)$ to a declarative sentence $S_j$. 
Each 4-tuple $\big(q_i, a_j, \prob{\vlm}(a_j), S_j \big)$ represents an evidence $\evidence{j}$. 

\paragraph{Check Evidence Reliability:}
Since \recoverr\ decides to make a prediction based on the collected evidences, the correctness likelihood of a given evidence is an upper bound for the correctness likelihood of the prediction. 
Since \recoverr aims to make predictions with correctness likelihood $\geq 1-r$, we only consider evidences $\evidence{j}$ that satisfy $\prob{\vlm}(\answer_j) \geq 1-r$ as \textit{reliable} for $\evidences{R}$. 
\textbf{\recoverr\ requires that the confidence estimate} $\prob{\vlm}(\answer)$ \textbf{is calibrated}. 
A confidence estimate being calibrated means that for all predictions whose confidence is $\alpha \in [0, 1]$, $\alpha$\% of the predictions are actually correct. 
In our experiments, we first evaluate calibration of our VLMs' confidence estimates (\S~\ref{subsubsec:vlm-calib}).

\paragraph{Check Evidence Relevance:} 
To decide whether a reliable evidence $\evidence{j}$ is also relevant, we adopt a defeasible reasoning approach~\cite{Rudinger2020ThinkingLA}. 
An evidence $\evidence{j}$ is considered relevant if its truth value affects the entailment probability of the hypothesis. 
We measure the absolute difference between the entailment probabilities of the hypothesis $\hypothesis$ conditioned on the evidence premise $S_j$ and its negated counterfactual i.e. $\bar{S}_j$.
\begin{align*}
    \rel(\evidence{j}) = \mid \prob{\nli}(\hypothesis | S_j) - \prob{\nli}(\hypothesis | \bar{S}_j) \mid
\end{align*}
Evidence $\evidence{j}$ is added to the relevant evidence set $\evidences{RR}$ if the relevance $\rel(\evidence{j})\geq\rel_{\texttt{min}}.$ 

\paragraph{Check Sufficiency of Collected Evidences:} 
After each round of evidence collection, we concatenate the $S_j$ for all evidences $\evidence{j} \in \evidences{RR}$ into a premise sentence $S_{RR}$ and calculate the hypothesis's entailment probability $\prob{\nli}(\hypothesis | S_{RR})$. 
If that probability meets $\prob{NLImin}$, we return prediction $\answer$, increasing coverage with sufficient estimated confidence that risk will not increase above $r$ as a result.
Else, after $N$ rounds, we abstain and return $\varnothing$.

\section{Experiments}
\label{sec:experiments}

\label{subsec:vlms}
\begin{figure*}[t]
    \centering
    \includegraphics[width=0.95\textwidth]{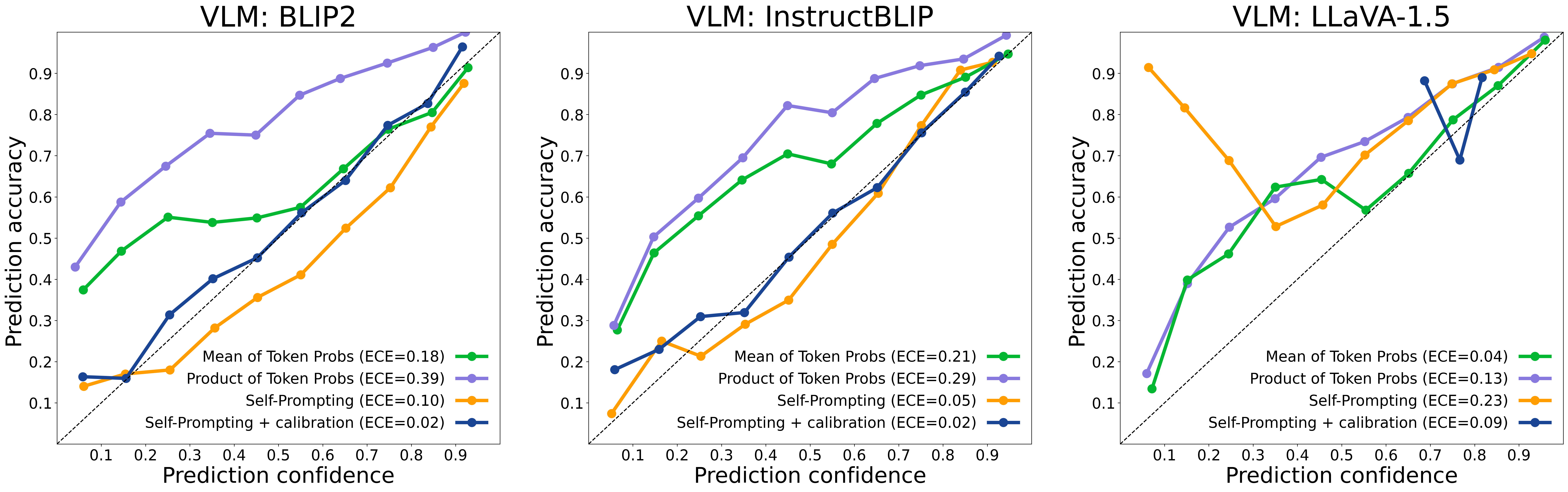}
    \caption{Calibration curves for BLIP-2, InstructBLIP and LLaVA-1.5 on A-OKVQA questions.}
    \label{fig:calibration}
\end{figure*}

We instantiate a selective prediction task using the A-OKVQA benchmark and compare \recoverr\ to simple threshold-based selective prediction with two different backbone VLMs with calibrated confidence estimates, using metrics that capture aspects and tradeoffs of risk and coverage.

\subsection{Vision-Language Reasoning Tasks: A-OKVQA and VQAv2}
\label{subsec:aokvqa}
A-OKVQA~\cite{schwenk2022okvqa} is a VQA task designed to require reasoning over external knowledge and commonsense alongside image-based information.
VQAv2~\cite{balanced_vqa_v2} is a VQA task that requires reasoning about images, sometimes involving commonsense reasoning. 
We evaluate on the A-OKVQA validation set ($n=1,075$) and 1,000 examples from the VQAv2 validation set. 
Both tasks involve open-ended answer generation, without any choices provided to the model.

For evaluating answer correctness, the standard VQA accuracy metric~\cite{Antol_2015_ICCV} has been shown to penalize correct VLM answers if they do not exactly match reference answers~\cite{agrawal2023reassessing}. 
Therefore, we use LAVE$_{\textsc{GPT-3.5}}$~\cite{manas2023improving} to measure the accuracy of predicted answers.
LAVE uses a large language model\footnote{We utilize the \texttt{gpt-3.5-turbo-16k-0613} checkpoint.} to estimate the semantic similarity of each predicted answer to the 10 crowdsourced answers in the benchmark.

\subsection{Vision-Language Models}

We experiment with three VLMs: BLIP2~\cite{li2023blip}, InstructBLIP~\cite{dai2023instructblip}, and LLaVA-1.5~\cite{liu2023improved}. Both BLIP models use FlanT5-XL as the LLM backbone.\footnote{\url{https://github.com/salesforce/LAVIS}} 
InstructBLIP and LLaVA-1.5 are instruction tuned on both VQAv2 and A-OKVQA; BLIP2 has not been trained on either task.

\subsubsection{Calibrated VLM Confidence Estimates}
\label{subsubsec:vlm-calib}

One key requirement for \recoverr is that, for a prediction $\answer$, the VLM can also return a \emph{calibrated} confidence score $\prob{\vlm}(\answer) \in [0, 1]$. 
Thus, a first step when applying \recoverr\ to a new VLM is identifying a confidence function that produces calibrated confidence estimates.

Confidence estimates for generative VLMs are not well defined.
Straightforward estimates such as the product of answer token likelihoods can severely underestimate model confidence due to factors like \emph{surface form competition}~\cite{holtzman2021surface}.
Inspired by~\citet{Tian2023JustAF}, we devise a Self-Prompting technique for estimating $\prob{\vlm}(\answer)$ by prompting the VLM to verify ``yes'' or ``no'' correctness of its predictions and examine the resulting probability distribution (full details in Appendix~\ref{sec:self-prompting}).
The Self-Prompting confidence can be further calibrated using Platt scaling~\cite{platt1999probabilistic}.

We compare three methods for extracting VLM confidence scores: product of token probabilities, mean of token probabilities, and Self-Prompting. 
For each VLM, we evaluate the calibration error of these confidence functions on a set of 5,000 examples from the A-OKVQA training data. 
The remaining 12,000 examples are used to calibrate Self-Prompting with Platt scaling.
We find that Self-Prompting (both \textit{off-the-shelf} and \emph{calibrated}) has the lowest calibration error for the BLIP models, whereas the mean of token probabilities has lowest error for LLaVA-1.5 (Figure~\ref{fig:calibration}). 
We therefore use these respective methods for estimating VLM confidence in our experiments with \recoverr.

\subsection{\recoverr\ Implementation Details}
\label{subsec:implementation}

\recoverr\ can be instantiated with different choices of models and system hyperparameters.

\begin{table*}[t]
    \centering
    \begin{tabular}{lllllllllll}
    \toprule
 &  & \multicolumn{4}{c}{Risk tolerance $r$ = 10\%} &  & \multicolumn{4}{c}{Risk tolerance $r$ = 20\%} \\ \cmidrule{3-6} \cmidrule{8-11}
\multicolumn{1}{c}{VLM} & \multicolumn{1}{c}{Method} & \risk ($\downarrow$) & \effrel ($\uparrow$) & \coverage ($\uparrow$) & \sprecall ($\uparrow$) &  & \risk ($\downarrow$) & \effrel ($\uparrow$) & \coverage ($\uparrow$) & \sprecall ($\uparrow$)  \\  \midrule
\multirow{3}{*}{\shortstack[l]{BLIP2\\(Off-the-shelf)}} & Vanilla SelPred & \phantom{0}6.1 & \phantom{0}3.4 & \phantom{0}3.8 & \phantom{0}6.0 &  & 14.9 & 17.1 & 24.1 & 34.2  \\ 
 & Vision Tools & \color{maroon}{13.8} & 17.4 & 23.5 & 33.5 &  & 17.0 & 23.5 & 35.1 & 48.3  \\ 
 & \recoverr & \redresult{14.3}{0.3} & \blueresult{18.8}{0.2} & \blueresult{26.0}{0.5} & \blueresult{37.1}{0.6} &  & \redresult{21.7}{0.8} & \blueresult{27.1}{0.8} & \blueresult{47.3}{0.3} & \blueresult{61.5}{0.8}  \\  \midrule
\multirow{3}{*}{\shortstack[l]{BLIP2\\(Calibrated)}} & Vanilla SelPred & \phantom{0}4.1 & \phantom{0}3.2 & \phantom{0}3.4 & \phantom{0}5.5 &  & 11.9 & 16.8 & 21.9 & 32.0  \\ 
 & Vision Tools & \color{maroon}{13.2} & 17.5 & 23.4 & 33.7 &  & 15.1 & 23.9 & 33.6 & 47.2  \\ 
 & \recoverr & \redresult{14.0}{0.3} & \result{17.3}{0.1} & \result{23.6}{0.2} & \result{33.6}{0.2} &  & \result{16.1}{0.7} & \blueresult{25.3}{0.5} & \blueresult{36.7}{0.2} & \blueresult{51.0}{0.6}  \\  \midrule
\multirow{3}{*}{\shortstack[l]{InstructBLIP\\(Off-the-shelf)}} & Vanilla SelPred & \phantom{0}9.3 & 20.5 & 25.1 & 34.8 &  & 17.2 & 38.3 & 57.7 & 72.2  \\ 
 & Vision Tools & \color{maroon}{10.5} & 26.1 & 32.9 & 44.8 &  & 17.6 & 39.6 & 60.5 & 75.3  \\ 
 & \recoverr & \redresult{10.9}{0.2} & \result{26.3}{0.3} & \result{33.5}{0.5} & \result{45.2}{0.6} &  & \result{17.9}{0.3} & \blueresult{41.3}{0.3} & \blueresult{63.7}{0.3} & \blueresult{78.9}{0.4}  \\ \midrule 
\multirow{3}{*}{\shortstack[l]{InstructBLIP\\(Calibrated)}} & Vanilla SelPred & \phantom{0}8.5 & 22.0 & 26.4 & 36.9 &  & 17.5 & 37.2 & 56.6 & 70.5  \\ 
 & Vision Tools & \color{maroon}{10.4} & 27.2 & 34.1 & 46.5 &  & 17.8 & 38.8 & 59.7 & 74.2  \\ 
 & \recoverr & \redresult{11.8}{0.7} & \blueresult{29.8}{0.8} & \blueresult{38.7}{1.3} & \blueresult{51.8}{1.4} &  & \result{18.6}{0.1} & \blueresult{41.4}{0.3} & \blueresult{65.0}{0.7} & \blueresult{79.8}{0.8}  \\ \midrule 
\multirow{3}{*}{\shortstack[l]{LLaVA-1.5}} & Vanilla SelPred & \phantom{0}8.2	& 21.5	& 25.1	& 33.0 &  & 16.5	& 40.1	& 59.3	& 69.9  \\ 
 & Vision Tools & \phantom{0}9.6	& 24.4	& 30.0	& 38.3 &  & 17.3	& 40.8	& 61.8	& 72.1  \\ 
 & \recoverr & \redresult{11.3}{}	& \blueresult{34.7}{}	& 4\blueresult{5.4}{} &	\blueresult{55.9}{} &  & 19.5	& \blueresult{44.6}{}	& \blueresult{72.7}{}	& \blueresult{81.9}{}  \\
     \bottomrule
    \end{tabular}
    \caption{Metric results as percentages on the A-OKVQA task at two risk tolerance levels.
    We evaluate selective prediction methods on the overall system risk (\risk), effective reliability (\effrel), coverage (\coverage) and recall (\sprecall). System risks in {\color{maroon}red} exceeded tolerance. 
    Measurements in {\color{navyblue}\textbf{blue}} indicate when \recoverr\ outperformed both baselines.}
    \label{tab:aokvqa_results}
\end{table*}

\subsubsection{\recoverr\ Model-based Tools}
\recoverr\ leverages a set of vision tools $\modelfamily{\vision}$ for extracting general visual information. 
In our experiments, the vision tools $\modelfamily{\vision}$ consists of LVIS~\cite{gupta2019lvis} for object detection, and Qwen-VL~\cite{bai2023qwen} for region captioning.
We use FlanT5-XL as a zero-shot sentence paraphraser $\model{\qas}$ and entailment model $\model{\nli}$. 
Since our VLMs use FlanT5-XL as the LLM backbone, our \recoverr\ instantiation does not require additional models. 
We use GPT-3.5 with temperature 1.0 for $\model{\qgen}$.
See Appendix~\ref{sec:llm_prompts} for prompts.

\subsubsection{\recoverr\ Hyperparameters}
We set $N=10$ rounds of evidence collection with $K=10$ evidences generated per turn. 
To count an evidence $\evidence{}$ as reliable, we set relevance threshold $\delta_{\texttt{min}}=0.2$. 
The minimum final hypothesis entailment confidence is $\prob{\texttt{NLImin}}=0.9$.

\subsection{Selective Prediction Baselines}
We compare \recoverr\ to a \textbf{Vanilla Selective Prediction} baseline~\cite{whitehead2022reliable}, where the model abstains if the VLM prediction confidence $g(\answer)=\{1\text{ if }\prob{\vlm}(\answer)\leq\gamma_{@r}\text{ else }0\}$, and to a \textbf{Vision Tools} baseline.
For \textbf{Vision Tools}, image caption from the VLM, objects detected by LVIS, and region captions from Qwen-VL are provided as evidence to the NLI model $\model{\nli}$ directly, with no rounds of evidence collection, equivalent to lines 1-5 of Algorithm~\ref{alg:recoverr-pseudocode} followed by lines 14-15.

\subsection{Evaluation Metrics}
We measure \textbf{risk} (\risk) and \textbf{coverage} (\coverage) (Eqs~\ref{eq:risk}, \ref{eq:coverage}) across system predictions and abstentions.
Additionally, we calculate \textbf{Effective Reliability} $\Phi_c$~\cite{whitehead2022reliable}, a metric that rewards correct predictions, assigns a penalty $c$ to incorrect predictions, and assigns zero reward to abstentions. 
We assign a penalty of $c=1$ for incorrect answers (\effrel). 
Finally, we define \textbf{Selective Prediction Recall} (\sprecall), the percentage of correct answers ($\text{Acc}(a) = 1$) in the dataset that were answered.
\begin{align*}
\text{\sprecall} & = \frac{\sum_{x_i \in \dataset} g(\answer_i) \cdot \mathbbm{1}\{ \text{Acc}(\answer_i) = 1 \}}{\sum_{x_i \in \dataset} \mathbbm{1}\{ \text{Acc}(\answer_i) = 1 \}}
\end{align*}


We note that the only stochastic component in our instantiation of \recoverr\ is the question generation model. 
We run three seeds for each \recoverr\ experiment and report average metric results. 
Table~\ref{tab:results-with-std} shows full results with standard deviation.

\section{Results and Analysis}
\label{sec:results}
We evaluate selective prediction methods on A-OKVQA at two specified risk tolerances: 10\% and 20\% risk. 
Table~\ref{tab:aokvqa_results} shows the results of \recoverr\ when applied to InstructBLIP and BLIP2, both with and without Platt scaling calibration. 

At 20\% risk tolerance, we find that \recoverr\ improves coverage, effective reliability, and recall over both baselines while staying at the specified risk tolerance. 
At 10\% risk tolerance, \recoverr\ and the Vision Tools baseline tend to slightly overshoot the risk tolerance (by 1-2\% for InstructBLIP and LLaVA-1.5, and 3-4\% for BLIP2). 
We further see that \recoverr\ shows large improvements of up to 20\% on coverage, 15\% on effective reliability and 33\% on recall for off-the-shelf BLIP2, calibrated InstructBLIP and LLaVA-1.5.

\begin{figure}[t]
    \centering
    \includegraphics[width=\columnwidth]{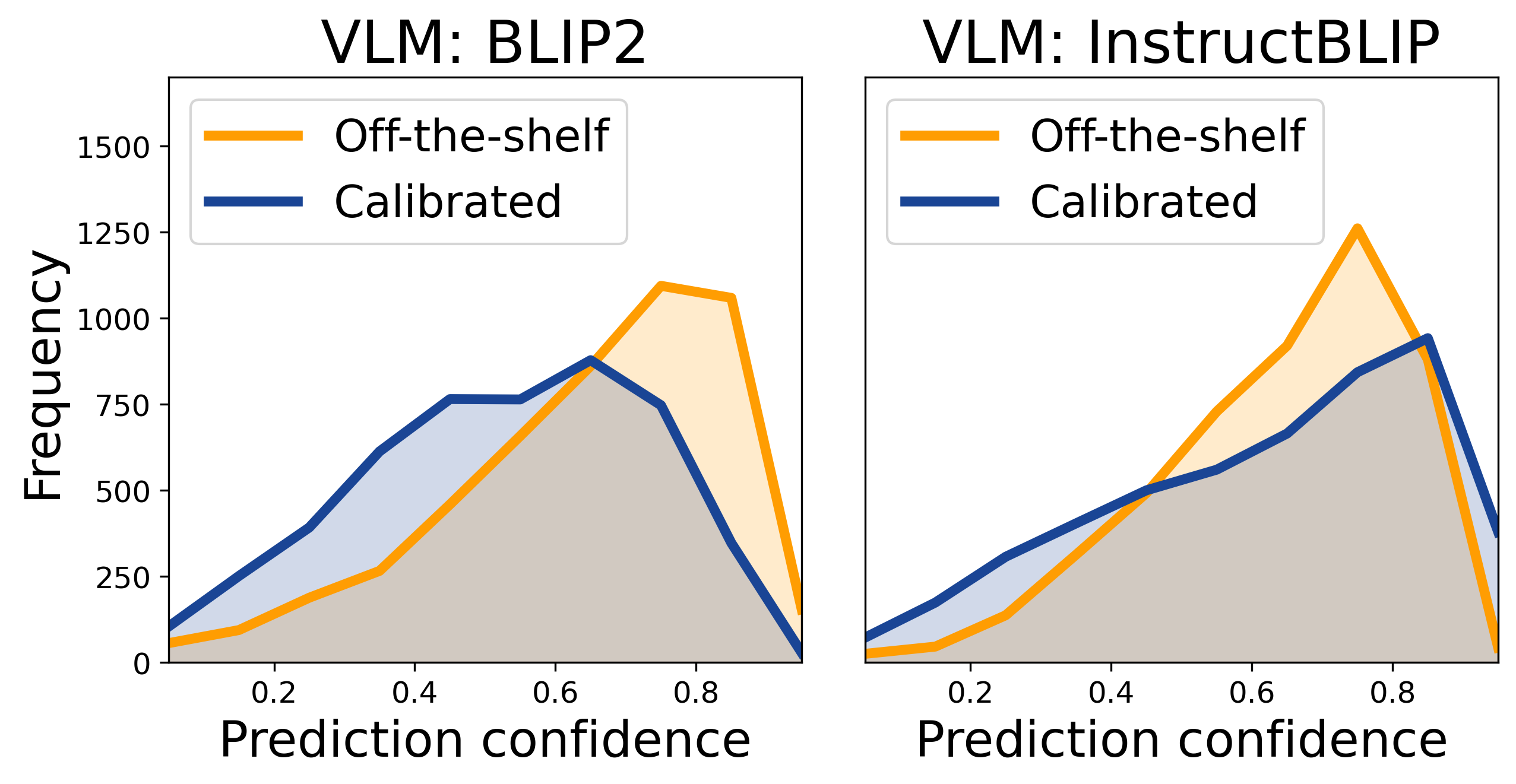}
    \caption{Distribution of VLM prediction confidences on A-OKVQA calibration set.}
    \label{fig:vlm-conf-distr}
\end{figure}
\begin{table}[t]
    \centering
    \small
    
    \begin{tabular}{l@{\hspace{0.1cm}}l@{\hspace{0.1cm}}c@{\hspace{0.2cm}}c@{\hspace{0.2cm}}c@{\hspace{0.2cm}}c}
    \toprule
\multicolumn{1}{c}{VLM} & \multicolumn{1}{c}{Method} & \risk ($\downarrow$) & \effrel ($\uparrow$) & \coverage ($\uparrow$) & \sprecall ($\uparrow$) \\  \midrule

\multirow{3}{*}{\shortstack[l]{BLIP2\\(Off-the-shelf)}} & Vanilla SelPred & \phantom{0}7.9	& 13.5	& 15.9	& 21.3 \\ 
 & Vision Tools &   \redresult{11.8}{} &	20.5 &	26.6 &	34.0  \\ 
 & \recoverr & \redresult{11.9}{} &	\blueresult{21.3}{}	& \blueresult{27.8}{} &	\blueresult{35.4}{}  \\  \midrule
 
\multirow{3}{*}{\shortstack[l]{BLIP2\\(Calibrated)}} & Vanilla SelPred & \phantom{0}7.7 & 13.1 & 15.5 & 20.9 \\ 
 & Vision Tools & \redresult{11.3}{} & 21.8 & 27.9 & 35.9 \\ 
 & \recoverr & \redresult{11.8}{} & \blueresult{22.9}{} & \blueresult{29.7}{} & \blueresult{38.1}{} \\  \midrule
 
\multirow{3}{*}{\shortstack[l]{InstructBLIP\\(Off-the-shelf)}} & Vanilla SelPred & 10.0 &	31.8	& 39.5 &	45.8 \\ 
 & Vision Tools & \redresult{10.8}{} &	34.8 &	43.1 &	50.2   \\ 
 & \recoverr & \redresult{10.7}{0.2} & 35.5	& \blueresult{45.0}{}	& \blueresult{51.9}{} \\ \midrule 
 
\multirow{3}{*}{\shortstack[l]{InstructBLIP\\(Calibrated)}} & Vanilla SelPred & 10.3&	32.3&	41.1&	47.5 \\ 
 & Vision Tools & \redresult{11.2}{}&	35.4 &	44.3 &	51.1 \\ 
 & \recoverr & \redresult{11.0}{}	&35.9 &	\blueresult{45.8}{}	& \blueresult{52.6}{} \\ \midrule 
 
\multirow{3}{*}{\shortstack[l]{LLaVA-1.5}} & Vanilla SelPred & 10.4 &	49.4 &	62.3 &	74.5 \\ 
 & Vision Tools & \color{maroon}{11.0} &	49.7 &	63.1 &	75.5  \\ 
 & \recoverr & \redresult{11.8}{} &	\blueresult{51.5}{} &	\blueresult{67.9}{} & 	\blueresult{79.2}{}   \\
     \bottomrule
    \end{tabular}
    \caption{Metric results as percentages on the VQAv2 task at a risk tolerance of 10\%.
    We evaluate selective prediction methods on overall system risk (\risk), effective reliability (\effrel), coverage (\coverage) and recall (\sprecall). System risks in {\color{maroon}red} exceeded tolerance. 
    Measurements in {\color{navyblue}\textbf{blue}} indicate when \recoverr\ outperformed both baselines.}
    \label{tab:vqav2_results}
\end{table}    

We highlight several takeaways (\textbf{T*}).
\textbf{T1}: For the same risk tolerance, vanilla selective prediction has lower coverage and recall for BLIP2 than InstructBLIP and LLaVA-1.5.
BLIP2 has not been trained on A-OKVQA, unlike the other two VLMs, and therefore produces more uncertain answers. We further see that \recoverr\ results in larger improvements for BLIP2.
\textbf{T2}: Off-the-shelf BLIP2 sees largest risk increase with \recoverr.
The calibration curves in Figure~\ref{fig:calibration} indicate that off-the-shelf BLIP2 makes overconfident estimates, leading to more unreliable evidences being collected and resulting in a large risk increase for \recoverr.
\textbf{T3}: At 10\% risk, most of the additional risk and recall is caused by Vision Tools, rather than \recoverr's evidence collection.
\textbf{T4}: At 20\% risk tolerance, \recoverr's evidence collection leads to greatest recall improvement for off-the-shelf BLIP2, calibrated InstructBLIP and LLaVA-1.5.

\textbf{T3\&4} can be understood by examining the confidence distribution of the VLMs' predictions (Figure~\ref{fig:vlm-conf-distr}). 
We see that off-the-shelf BLIP2 has more predictions with greater than 80\% confidence compared to calibrated BLIP2, with the trend reversed for InstructBLIP. 
This means that \recoverr\ finds more reliable evidences ($\prob{\vlm}(\answer) \geq 1-r$) with off-the-shelf BLIP2 and calibrated InstructBLIP, resulting in higher recall for \recoverr\ with those two VLMs, thus explaining \textbf{T3\&4}. 
These findings indicate that \recoverr\ works best when the VLM is \emph{strongly confident} on correct answers.

We also experiment on VQAv2, at 10\% risk tolerance.\footnote{Since the VLMs have strong performance on VQAv2, we do not perform experiment at 20\% risk tolerance.} 
In Table~\ref{tab:vqav2_results}, we see that \recoverr\ improves system reliability and recovers more correct answers than the baselines, while crossing the risk threshold by only 1--2\%. 
However, the improvements are smaller compared to the more complex A-OKVQA task, indicating that \recoverr\ may be more useful for complex reasoning tasks. 
We also observe that, similar to A-OKVQA, the largest improvements are for BLIP2, which wasn't finetuned on the VQAv2 task.

\subsection{Reliability and Relevance Ablations}

We perform an ablation of the reliability and relevance components of \recoverr\, for calibrated BLIP2 and InstructBLIP, at a risk tolerance of 20\%. 
In the \recoverr\ formulation, for an evidence $\evidence{j}$ to be considered reliable, the VLM confidence $\prob{\vlm}(\answer_j | I, \question_j)$ must be at least $1-r$, and for it to be considered relevant, the defeasible relevance $\rel(\evidence{j})$ must be at least $\rel_{\texttt{min}}=0.2$. 
For the reliability ablation, we lower the minimum evidence confidence to be $0.5$ instead of $1-r=0.8$. 
For the relevance ablation, we set $\rel_{\texttt{min}}=0$. 
\label{subsec:ablation}
\begin{table}[t]
\centering
\begin{tabular}{llll} 
\toprule
Method & \risk ($\downarrow$) & \effrel ($\uparrow$)  & \sprecall ($\uparrow$) \\
  \midrule
\multicolumn{4}{c}{VLM : Calibrated BLIP2} \\
\midrule
    \recoverr\  & 16.1  & 25.3 &  51.0 \\
     \quad - Reliability & \redresult{28.4} &  & 31.6 &  85.9\\
     \quad - Relevance & 16.1 & 25.3  & 51.1\\
\midrule
\multicolumn{4}{c}{VLM : Calibrated InstructBLIP} \\
\midrule
    \recoverr\  & 18.6 &  41.4  & 79.8\\
     \quad - Reliability & \redresult{25.9} & &  40.5 & 91.9 \\
     \quad - Relevance & 18.6 & \redresult{39.7} & & \redresult{76.9} &\\
\bottomrule
\end{tabular}
\caption{Ablation of the reliability and relevance components of \recoverr, at risk tolerance of $20\%$.}
\label{tab:ablation}
\end{table}
\begin{table}[t]
    \centering
    \begin{tabular}{llll}
    \toprule
         \multicolumn{1}{c}{Calibrated VLM} & \multicolumn{1}{c}{Method} & \risk ($\downarrow$) &  \sprecall ($\uparrow$)  \\
         \midrule
         \multicolumn{4}{c}{Task: OK-VQA (20\% risk tolerance)} \\
         \midrule
         \multirow{2}{*}{BLIP2} & Vanilla & 15.4 & 25.4 \\
         & \recoverr &  \redresult{25.7} & & 45.0 \\
         \cmidrule{2-4}
         \multirow{2}{*}{InstructBLIP} & Vanilla & 18.5 & 56.3 \\
         & \recoverr & \redresult{22.8} & & 70.8 \\
         \midrule
         \multicolumn{4}{c}{Task: Sherlock (10\% risk tolerance)} \\
         \midrule
         \multirow{2}{*}{BLIP2} & Vanilla & 11.1	& 24.9 \\
         & \recoverr & \redresult{15.7} & & 25.8  \\
         \cmidrule{2-4}
         \multirow{2}{*}{InstructBLIP} & Vanilla & \phantom{0}9.7 & 33.4 \\
         & \recoverr & 10.9 & 35.5 \\
         \bottomrule
    \end{tabular}
    \caption{\recoverr using a VLM calibrated for A-OKVQA can be directly applied to new tasks.}
    \label{tab:new-tasks}
\end{table}
\begin{table}[t]
    \centering
    \begin{tabular}{lllll}
    \toprule
         \multicolumn{1}{c}{$\model{\qgen}$} & \risk ($\downarrow$) & \effrel ($\uparrow$) & \coverage ($\uparrow$) & \sprecall ($\uparrow$)  \\
         \midrule
         \multicolumn{5}{c}{VLM : Calibrated BLIP2} \\
         \midrule 
         ChatGPT & 16.1 & 25.3 & 36.7 & 51.0 \\
         Mistral-7B & 16.9 & 25.4 & 37.9 & 52.2 \\
         Tulu-2-7B & 15.8 & 24.8 & 35.8 & 50.0 \\
         \midrule
         \multicolumn{5}{c}{VLM : Calibrated InstructBLIP} \\
         \midrule 
         ChatGPT & 18.6 & 41.4 & 65.0 & 79.8 \\
         Mistral-7B & 19.1 & 39.9 & 63.8 & 77.8 \\
         Tulu-2-7B & 18.0 & 40.9 & 63.3 & 78.2 \\
         \bottomrule
    \end{tabular}
    \caption{Effect of different question generation models on \recoverr\ performance, at 20\% risk tolerance.}
    \label{tab:qgen-models}
\end{table}
\begin{table*}[t]
    \centering
    \small
    \begin{tabular}{p{0.16\textwidth} p{0.27\textwidth}  p{0.4\textwidth}}
    \toprule
    & & \multicolumn{1}{c}{Evidences $\evidences{RR}$ collected by \recoverr} \\
    \multicolumn{1}{c}{Image} & \multicolumn{1}{c}{Initial prediction} & \qualexample{\{LLM Question $q_j$\}}{\{VLM Answer $\answer_j$\}}{$\prob{\vlm}(\answer_j)$} \\
    \midrule
     \multirow{1}{*}{\frame{\includegraphics[width=0.16\textwidth]{figs/vegetarian-meal.png}}} & 
     \textbf{Question:} What kind of a person usually eats food like this? \newline \textbf{G.T. answers:} vegetarian, vegan, healthy \newline \textbf{VLM:} vegetarian (conf=0.57) & 
     \qualexample{Does the image depict a variety of plant-based ingredients?}{yes}{0.95} \newline \qualexample{Does the food predominantly consist of fruits and vegetables?}{yes}{0.80} \newline 
     \qualexample{Are there any meat items in the image?}{no}{0.86} \vspace{0.4em} \newline \textbf{Result: Correctly predicted} \\ 
     \midrule
     \multirow{1}{*}{\frame{\includegraphics[width=0.16\textwidth]{figs/bus.png}}} & 
     \textbf{Question:} The colors on the bus match the colors on what flag? \newline \textbf{G.T. answers:} Sweden, Ukraine \newline \textbf{VLM:} U.K. (conf=0.53) & 
     \qualexample{Is there any blue color on the bus?}{yes}{0.95} \newline \qualexample{What is the color of the bus?}{yellow and blue}{0.93} \newline \vspace{0.2em} \newline \textbf{Result: Correctly abstained}  \\ 
    \bottomrule
    \end{tabular}
    \caption{Examples from the A-OKVQA task where, given a low-confidence VLM prediction, \recoverr\ collects a set of highly-confident evidences $\evidences{RR}$ that corroborate (example 1) or contradict (example 2) the prediction.}
    \label{tab:qualitative}
\end{table*}

Table~\ref{tab:ablation} shows the results of the reliability and relevance ablation. 
We see that loosening the evidence confidence bound to be $0.5$ instead of $1-r$ causes a marked increase in risk for both VLMs, with system risk crossing the risk tolerance significantly. 
Ablating the relevance component results in a decrease in recall and effective reliability for InstructBLIP, but virtually no effect for BLIP2.

\subsection{Question Generation Model Ablations}
\label{subsec:qgen}

The specific instantiation of \recoverr\ involves a choice of LLM for the question generation model, $\model{\qgen}$. 
While our experiments so far have used GPT-3.5, we demonstrate that \recoverr\ works similarly well with open-source LLMs as well. 
We compare \recoverr\ performance with GPT-3.5 against two open-source LLMs: Mistral-7B-Instruct~\cite{jiang2023mistral} and Tulu-2-7B~\cite{ivison2023camels}. 
Our results in Table~\ref{tab:qgen-models} demonstrate that the specific instantiation of $\model{\qgen}$ did not make a significant difference to \recoverr's performance, for both calibrated BLIP2 and InstructBLIP.

\subsection{Generalizing to New Tasks}
\label{subsec:task-generalization}
We examine whether our instantiation of \recoverr calibrated for the A-OKVQA task can be directly applied to new tasks without additional tuning (Table~\ref{tab:new-tasks}). 
When we apply \recoverr\ with A-OKVQA-calibrated VLMs to OK-VQA~\cite{marino2019ok} at 20\% risk tolerance, we observe strong improvements in recall (15-20\%), but also increased risk (2-5\% above the risk tolerance). 
Applying \recoverr\ to Sherlock~\cite{hesselhwang2022abduction}, an abductive reasoning task, at 10\% risk we observe smaller improvements in recall. 
These results indicate that \recoverr\ requires some degree of task-specific tuning. 
Appendix~\ref{sec:new-tasks} contains full task details.

\subsection{Qualitative \recoverr\ Examples}
We present some examples of evidences collected by \recoverr\ in Table~\ref{tab:qualitative}. 
The collected evidences $\evidences{RR}$ can be used to either corroborate correct VLM predictions, or contradict incorrect ones. 
In example 1,  when the VLM predicts that a vegetarian is likely to eat this meal, \recoverr\ collects supporting evidences indicating that the food does not contain any meat products.
In example 2, when the VLM predicts that the colors of the bus match those of the U.K. flag, \recoverr\ finds that the color of the bus is highly likely to be yellow and blue, thus contradicting the VLM's prediction (since the U.K. flag does not contain yellow).

\section{Conclusion and Related Works}
\label{sec:rw}

We introduce \recoverr, an algorithm to improve the coverage of a selective VLM system while respecting a user-specified risk tolerance level.
\recoverr\ verifies low-confidence VLM predictions by recovering high-confidence evidences in the image that support the prediction. 
We instantiate a selective prediction task on the A-OKVQA reasoning benchmark and demonstrate the quantitative advantages of \recoverr\ for inference-time selective prediction that holds across different backbone VLMs, choice of underlying LLM generators, and even gains some ground when transferred to different benchmarks without recalibration.

A large breadth of prior work has studied the ability of models to abstain from answering~\cite{chow1957optimum, de2000reject, el2010foundations}. This ability is especially important when the model's prediction cannot be trusted~\cite{jiang2018trust}, particularly for OOD~\cite{DBLP:conf/acl/KamathJL20} and adversarial~\cite{varshney2022investigating} inputs. 
Similar to our work, low coverage caused by low risk tolerance has also been observed and mitigated in text-only selective prediction systems~\cite{Varshney2023PostAbstentionTR}.

In the multimodal domain, previous works have explored abstention for error recovery~\cite{DBLP:journals/corr/abs-2312-08366, khan2023selective}. 
\citet{DBLP:conf/eccv/WhiteheadPS0DRR22} establish the ReliableVQA framework, wherein VQA models have the option to abstain from answering. 
\citet{DBLP:journals/corr/abs-2310-00647} evaluates VQA models on their ability to abstain when a question does not apply to the image. 
\citet{DBLP:conf/cvpr/DancetteWMVSCCR23} trains VQA models to abstain for out-of-distribution inputs. 
In contrast to \emph{learning to abstain}, we \emph{reduce abstention} of multimodal selective prediction while maintaining reliability.

Using LLMs to query a pool of vision experts, either through program generation~\cite{gupta2023visual, surismenon2023vipergpt, subramanian2023modular}, or iterative querying~\cite{zeng2022socratic, shen2023hugginggpt, you2023idealgpt, yang2023mm}, can decompose vision-language reasoning.
We use LLMs to query an image for visual evidence, similar to RepARe~\cite{prasad2023rephrase} which extracts additional visual context to rephrase underspecified VQA questions. 

\section*{Limitations}
The plug-and-play capability of \recoverr, while facilitating adaptability, does introduce a degree of engineering complexity due to the sequential dependencies in predictions. 
While it opens up flexible choices in modules, it also introduces an overhead on response time.
Our primary method black-box APIs, and we also tested it with open-source models, as detailed in the Table \ref{tab:qgen-models}. 
We leave conducting comprehensive testing with various open-source models for future work.

The curated benchmarks tailored for VLM evaluation, while being beneficial, may necessitate further exploration into scenarios that better align with ecological validity \cite{DBLP:journals/corr/abs-2007-14435} and real-world applications for abstaining. 
Notably, our focus on image understanding excludes considerations of underspecified, ambiguous, or safety-related abstaining, opening avenues for expanding the scope of our approach in the future.

Our experiments are restricted to English, a consequence of leveraging pretrained VLMs and LLMs, offers opportunities for future advancements in multilingual contexts. 

Finally, we assume that the base VLMs are calibrated for confidence estimation. Although this is effective to some extent, it leaves room for future exploration into alternative methodologies for making the models well-calibrated.

\section{Acknowledgements}
Jack Hessel partially completed this work while at AI2.

\clearpage

\bibliography{custom}

\clearpage

\appendix

\section{Self-Prompting for Producing Calibrated VLM Confidence Estimates}
\label{sec:self-prompting}

In order to do selective prediction, it is important that we use a calibrated confidence estimate \textit{i.e.} the model confidence $g(x)$ reflects the true probability of its output being correct~\cite{platt1999probabilistic}. 
The calibration of a model's confidence measure is typically evaluated using the Expected Calibration Error~\cite{guo2017calibration} over a calibration set. 

Calibration has typically been studied in discriminative models, using the probability of the predicted class as the confidence estimate. 
However, it is unclear what makes for a good confidence estimate for a generative VLM that produces an answer $a$ with multiple tokens.
One solution is to sum or average all the log probabilities in the answer token sequence string; however, this will be an underestimate of the model's confidence due to surface form competition~\cite{holtzman2021surface}.

Instead, we experiment with a \emph{self-prompting}, approach where the VLM is prompted to verify the correctness of its answer. 
Specifically, after predicting an answer $\answer$ from the VLM for the question $\question$, we present the following prompt \texttt{prompt($\question, \answer$)} to the VLM:
\begin{quote}
    Question: \{$\question$\}\\
    Answer: \{$\answer$\}\\
    Is the given answer correct for the question? Answer yes or no:
\end{quote}
We look at the VLM's next token probability distribution $P(\cdot | \texttt{prompt}, I)$, and compute confidence estimate by looking at the next token probabilities for the \texttt{yes} and \texttt{no} tokens:
\begin{align*}
    P_{\vlm}(\answer | \question) = \frac{P(\texttt{yes} | \texttt{prompt}, I)}{P(\texttt{yes} | \texttt{prompt}, I) + P(\texttt{no} | \texttt{prompt}, I)}
\end{align*}

\subsection{Platt Scaling}
Platt scaling~\cite{platt1999probabilistic} is a technique for calibrating classifiers by training a logistic regression model over the logits for the output classes. 
To train a Platt scaling calibrator, for 12,000 examples in the A-OKVQA training data, we use Self-Prompting as described above for the off-the-shelf VLMs to extract logits for the $\texttt{yes}$ and $\texttt{no}$ tokens. 
We train a logistic regression model over these logits to predict whether the VLM prediction was correct or not. 
Figure~\ref{fig:calibration} shows the effect of Platt scaling calibration on the VLM confidence estimates.
\section{Prompts for Text-Only Language Models}
\label{sec:llm_prompts}

Table~\ref{tab:llm_prompts} contains the prompts we used for the various language-only functions of \recoverr.

\begin{table*}[t]
    \centering
    \small
    \begin{tabular}{p{0.15\textwidth} p{0.05\textwidth}  p{0.2\textwidth} p{0.55\textwidth}}
    \toprule
         \multicolumn{1}{c}{Function} & \multicolumn{1}{c}{Model} & \multicolumn{1}{c}{Inputs} & \multicolumn{1}{c}{Prompt}  \\
    \midrule
        Question generation & $\model{\qgen}$ & $\question$: The question the VLM is trying to confidently answer \newline $\answer$:The VLM prediction for above question \newline $\evidences{R} = [\evidence{1}, \evidence{2}, ...]$: A list of reliable evidences that have been collected about the image so far \newline $K$: The number of questions to generate in this turn. &  You are an AI assistant who has rich visual commonsense knowledge and strong reasoning abilities. You will be provided with:
\newline 1. A target question about an image that you are trying to answer.
\newline 2. Although you won't be able to directly view the image, you will receive a general caption that might not be entirely precise but will provide an overall description.
\newline 3. You may receive some additional evidences about the image.
\newline Your goal is: To effectively analyze the image and select the correct answer for the question, you should break down the main question into several sub-questions that address the key aspects of the image.
\newline
\newline
What you already know about the image:
\newline $\evidence{1}$
\newline $\evidence{2}$
\newline ...
\newline 
\newline Target question: $\{ \question \}$. Generate $K$ sub-questions that might help you confirm whether the answer to the target question is '$\{ \answer \}$'. 
\newline Here are the rules you should follow when listing the sub-questions:
\newline 1. Ensure that each sub-question is independent. It means the latter sub-questions shouldn't mention previous sub-questions.
\newline 2. The sub-questions should be separated by a newline character.
\newline 3. Each sub-question should start with "What" or "Is".
\newline 4. Each sub-question should be short (less than 10 words) and easy to understand.
\newline 5. The sub-question are necessary to distinguish the correct answer.
 \\
        \midrule
        Sentence paraphrasing & $\model{\qas}$ & $\question$, $\answer$: The question-answer pair that needs to be paraphrased into a sentence & 
        Rephrase the question and answer into a single statement. \newline The re-phrased statement should summarize the question and answer. \newline The re-phrased statement should not be a question. 
         \newline \newline Question: Is the dog herding or guiding the cows?\newline  Answer: guiding \newline  Statement: The dog is guiding the cows.
        \newline \newline Question: Are there any other written numbers visible in the image? \newline Answer: no \newline Statement: There are no other written numbers visible in the image.
        \newline \newline Question: Which color of clothing is unique to just one of the two people here? \newline Answer: black \newline Statement: The color of clothing that is unique to just one of the two people here is black.
        \newline \newline Question: Does the picture on the screen involve any human subjects or animals? \newline Answer: human \newline Statement: The picture on the screen involves human subjects.
        \newline \newline Question: $\{ \question \}$ \newline Answer: $\{\answer \}$ \newline Statement: 
            \\
        \midrule Checking evidence sufficiency & $\model{\nli}$ & $\hypothesis$: Hypothesis statement \newline $\evidences{RR}$: A list of reliable and relevant evidences [$\evidence{1}, \evidence{2}, ...$] & Premise: $\{ \text{evidence statements in }\evidences{RR}\text{ concatenated together}\}$ \newline \newline Hypothesis: $\{\hypothesis\}$ \newline Can we infer the hypothesis from the premise? Options: yes, no. Answer:  \\
    \bottomrule
    \end{tabular}
    \caption{Prompts for text-only language models}
    \label{tab:llm_prompts}
\end{table*}
\section{OK-VQA and Sherlock Task Details}
\label{sec:new-tasks}
We apply \recoverr\ using A-OKVQA-calibrated VLMs for two new tasks, OK-VQA and Sherlock.

OK-VQA~\cite{marino2019ok} is a VQA task that, similar to A-OKVQA, requires external knowledge and commonsense reasoning. 
We evaluate on the OK-VQA validation set, consisting of 5,046 instances.

Sherlock~\cite{hesselhwang2022abduction} is an abductive reasoning task, where images and image regions are paired with inferences. 
The original task formulations in Sherlock (retrieval, localization, human rating comparison) do not directly apply to generative VLMs that take image and text inputs and produce text outputs. 
We re-formulate the comparison task into a binary judgment task, where the VLM must output whether a given inference is true for an image or not. 
An image-region-inference triple is paired with two human ratings that rate the inference as true, false, or neutral. 
We treat instances where both humans agree that the inference is true as positive instances, and both humans agree that it is false as negative instances. 
We end up with a dataset of 561 image-region-inference triples. 
Each triple is presented to the VLM by drawing a region box onto the image, and forming a templated question in the following format:
\begin{quote}
    Is the given inference true for the given image or not? \{ Inference \} Options: yes, no
\end{quote}

\begin{table*}[t]
    \centering
    \small
    \begin{tabular}{lllllllllll}
    \toprule
 &  & \multicolumn{4}{c}{Risk tolerance $r$ = 10\%} &  & \multicolumn{4}{c}{Risk tolerance $r$ = 20\%} \\ \cmidrule{3-6} \cmidrule{8-11}
\multicolumn{1}{c}{VLM} & \multicolumn{1}{c}{Method} & \risk ($\downarrow$) & \effrel ($\uparrow$) & \coverage ($\uparrow$) & \sprecall ($\uparrow$) &  & \risk ($\downarrow$) & \effrel ($\uparrow$) & \coverage ($\uparrow$) & \sprecall ($\uparrow$)  \\  \midrule
\multirow{3}{*}{\shortstack[l]{BLIP2\\(Off-the-shelf)}} & Vanilla SelPred & \phantom{0}6.1 & \phantom{0}3.4 & \phantom{0}3.8 & \phantom{0}6.0 &  & 14.9 & 17.1 & 24.1 & 34.2  \\ 
 & Vision Tools & \color{maroon}{13.8} & 17.4 & 23.5 & 33.5 &  & 17.0 & 23.5 & 35.1 & 48.3  \\ 
 & \recoverr & \redresult{14.3}{0.3} & \blueresultwithstdev{18.8}{0.2} & \blueresultwithstdev{26.0}{0.5} & \blueresultwithstdev{37.1}{0.6} &  & \redresult{21.7}{0.8} & \blueresultwithstdev{27.1}{0.8} & \blueresultwithstdev{47.3}{0.3} & \blueresultwithstdev{61.5}{0.8}  \\  \midrule
\multirow{3}{*}{\shortstack[l]{BLIP2\\(Calibrated)}} & Vanilla SelPred & \phantom{0}4.1 & \phantom{0}3.2 & \phantom{0}3.4 & \phantom{0}5.5 &  & 11.9 & 16.8 & 21.9 & 32.0  \\ 
 & Vision Tools & \color{maroon}{13.2} & 17.5 & 23.4 & 33.7 &  & 15.1 & 23.9 & 33.6 & 47.2  \\ 
 & \recoverr & \redresult{14.0}{0.3} & \result{17.3}{0.1} & \result{23.6}{0.2} & \result{33.6}{0.2} &  & \result{16.1}{0.7} & \blueresultwithstdev{25.3}{0.5} & \blueresultwithstdev{36.7}{0.2} & \blueresultwithstdev{51.0}{0.6}  \\  \midrule
\multirow{3}{*}{\shortstack[l]{InstructBLIP\\(Off-the-shelf)}} & Vanilla SelPred & \phantom{0}9.3 & 20.5 & 25.1 & 34.8 &  & 17.2 & 38.3 & 57.7 & 72.2  \\ 
 & Vision Tools & \color{maroon}{10.5} & 26.1 & 32.9 & 44.8 &  & 17.6 & 39.6 & 60.5 & 75.3  \\ 
 & \recoverr & \redresult{10.9}{0.2} & \result{26.3}{0.3} & \result{33.5}{0.5} & \result{45.2}{0.6} &  & \result{17.9}{0.3} & \blueresultwithstdev{41.3}{0.3} & \blueresultwithstdev{63.7}{0.3} & \blueresultwithstdev{78.9}{0.4}  \\ \midrule 
\multirow{3}{*}{\shortstack[l]{InstructBLIP\\(Calibrated)}} & Vanilla SelPred & \phantom{0}8.5 & 22.0 & 26.4 & 36.9 &  & 17.5 & 37.2 & 56.6 & 70.5  \\ 
 & Vision Tools & \color{maroon}{10.4} & 27.2 & 34.1 & 46.5 &  & 17.8 & 38.8 & 59.7 & 74.2  \\ 
 & \recoverr & \redresult{11.8}{0.7} & \blueresultwithstdev{29.8}{0.8} & \blueresultwithstdev{38.7}{1.3} & \blueresultwithstdev{51.8}{1.4} &  & \result{18.6}{0.1} & \blueresultwithstdev{41.4}{0.3} & \blueresultwithstdev{65.0}{0.7} & \blueresultwithstdev{79.8}{0.8}  \\
     \bottomrule
    \end{tabular}
    \caption{Metric results as percentages on the A-OKVQA task at two risk tolerance levels.
    We evaluate selective prediction methods on the overall system risk (\risk), effective reliability (\effrel), coverage (\coverage) and recall (\sprecall). System risks in {\color{maroon}red} exceeded tolerance. 
    Measurements in {\color{navyblue}\textbf{blue}} indicate when \recoverr\ outperformed both baselines.}
    \label{tab:results-with-std}
\end{table*}
\end{document}